\documentclass[letterpaper, 10 pt, conference]{ieeeconf}  

\IEEEoverridecommandlockouts                              

\usepackage{graphicx}
\usepackage{subfigure}
\usepackage{amsmath}
\usepackage{amssymb}
\usepackage{bm}
\usepackage{color}
\usepackage{caption}
\usepackage{algorithm}
\usepackage{algorithmic}

\title{\LARGE \bf
Human Following for Wheeled Robot with Monocular Pan-tilt Camera
}

\author{Zheng Zhu, Hongxuan Ma, Wei Zou
\thanks{Zheng Zhu, Hongxuan Ma, Wei Zou are with Institute of Automation, Chinese Academy of Sciences and University of Chinese Academy of Sciences.}
}

\begin{document}

\maketitle
\thispagestyle{empty}
\pagestyle{empty}


\begin{abstract}
Human following on mobile robots has witnessed significant advances due to its potentials for real-world applications. Currently most human following systems are equipped with depth sensors to obtain distance information between human and robot, which suffer from the perception requirements and noises. In this paper, we design a wheeled mobile robot system with monocular pan-tilt camera to follow human, which can stay the target in the field of view and keep following simultaneously. The system consists of fast human detector, real-time and accurate visual tracker, and unified controller for mobile robot and pan-tilt camera. In visual tracking algorithm, both Siamese networks and optical flow information are exploited to locate and regress human simultaneously. In order in perform following with a monocular camera, the constraint of human height is introduced to design the controller.
In experiments, human following are conducted and analysed in simulations and a real robot platform, which demonstrate the effectiveness and robustness of the overall system.


\end{abstract}

\section{Introduction}
Tracking and following human on mobile robots within cameras are of significance for human-computer interaction \cite{luo2019end, wang2017motion}, automatic driving \cite{verma2018vehicle,buyval2018realtime}, personal assistant robot \cite{hirose2015personal,zhang2019exploiting,li2019state} and service robot \cite{bajcsy2019scalable,bellotto2009multisensor, yao2017monocular,maselection,kang2019adaptive}. The core technology for the following robot mainly consists of human visual tracker and robot controller. The former tracks a specified human in a changing video sequences automatically given a detected bounding box in the first frame, while the controller generates necessary motion commands so that the robot can follow the target human.

The core problem of visual tracking is how to detect and locate the object accurately and fast in human following scenarios with occlusions, shape deformation, illumination variations \cite{liang2018planar,OTB2013, dressel2019hunting, OTB2015, VOT2017, VOT2018}. Similarly with other computer vision tasks \cite{AlexNet, ResNet, li2019attention, FPN, zhang2019fastpose, zhu2018end,DeepFace, FaceNet,zhu2018two, FasterRCNN,zhu2018action, zhu2019convolutional},
deep convolutional networks have achieved favourable performance in recent tracking benchmarks \cite{wang2018deep,DLT,Deeptrack,DeepSRDCF, HCF,HDT,VITAL,CFNet, CREST,PTAV,MDNet, CCOT, ECO, FlowTrack, UCT, SiamFC, SiamRPN, DaSiamRPN, bai2018multi}. In this paper, both Siamese networks \cite{SiamRPN, DaSiamRPN} and optical flow information \cite{FlowTrack,huang2018optical,huang2018efficient} are exploited to locate and regress human simultaneously. Besides, negative pairs of human are emphasised to suppress the response of distractor. A simple yet effective failure recovery strategy is proposed to handle the occlusion and out-of-view during tracking.

Since distance information of target is essential for robot perception and control, depth sensors are always equipped for following task \cite{razlaw2019detection,gulalkari2014object,zhu2016std,gulalkari2015object,mohamed2016stereo,gupta2017novel,IJRA, kobilarov2006people}. Laser scans with camera are always used to detect and track the fixed height legs of human \cite{kobilarov2006people}, which cannot provide robust features for discriminating the different persons following task.  Kinect cameras are frequently adopted in the robotics community \cite{gulalkari2014object, gulalkari2015object, gupta2017novel}, whose minimum distance requirement and sensitivity to the illumination variations limit its applications. Some robots \cite{mohamed2016stereo} are equipped with stereo vision systems to reconstruct the depth information, which suffer from baseline configuration, camera calibration and field of view problems.
In paper \cite{wang2018accurate}, monocular camera with an ultrasonic sensor are adopted to implement the human tracking system. However, the accuracy of ultrasonic sensor is always effected by reflection problem and noise. Above sensors are fixed to the robots, which limits the range of perception. In this paper, we develop a wheeled mobile robot system with monocular pan-tilt camera, which does not need the distance information provided by depth sensors. Besides, The camera can actively track human using pan-tilt motors.
In order in perform following with a monocular camera, the constraint of human height is introduced to design unified controller for wheeled robot and pan-tilt camera.
%
%

The rest of this paper is organized as follows:  Section \uppercase\expandafter{\romannumeral2} describes our designed human following systems, including robot configuration, fast human detector, real-time and accurate visual tracker, and unified controller for mobile robot within pan-tilt camera. Section \uppercase\expandafter{\romannumeral3} shows experiment results on human following scenarios. Section \uppercase\expandafter{\romannumeral4} concludes the paper with a summary.



\section{Following Human Utilizing Mobile Robot within Monocular Pan-tilt Camera}

In this section, we introduce the overall framework of human following system at first. Then the separate parts, including robot system and coordinate definition, fast human detector, real-time and accurate visual tracker, and unified controller for mobile robot within pan-tilt camera are detailed described respectively.

\subsection{Overall Human Following Framework}

\begin{figure*}[htbp]
  \centering
  \includegraphics[width=0.8\linewidth]{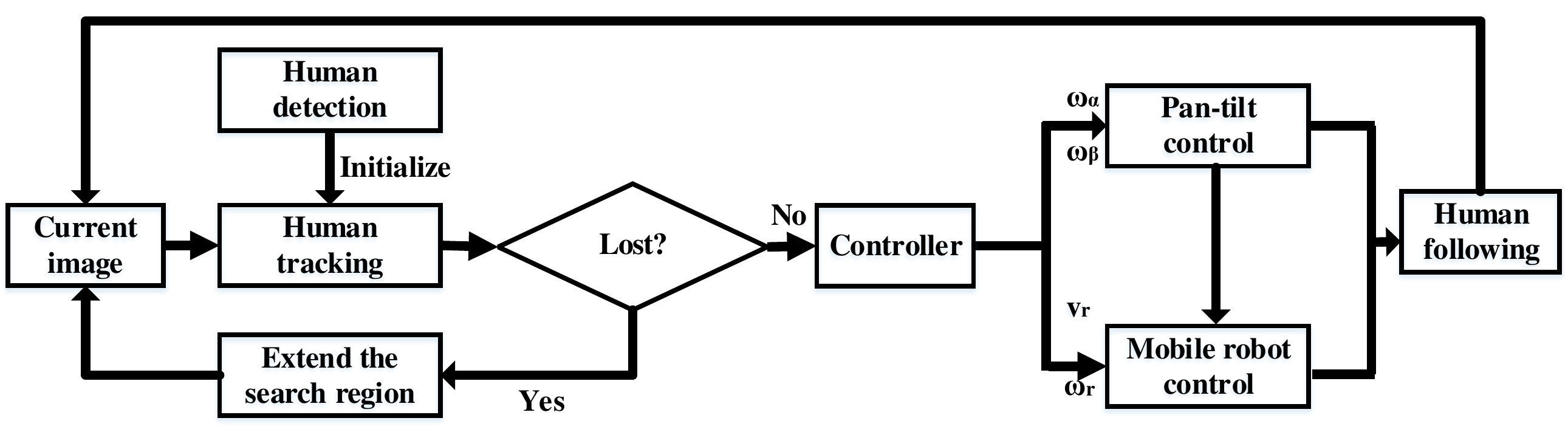}
  \caption{\small The overall framework for human following.}
  \label{overall_framework}
\end{figure*}

The overall human following framework consists of human detection and tracking in captured video streaming and controller for mobile robot within pan-tilt camera. At the beginning of human following, the human bounding box is indicated by tiny YOLO \cite{YOLO} detector. Once the initial box is given, human tracking loop is performed using the proposed \emph{FlowTrack++} algorithm. Tracking results are fed into the controller for mobile robot and pan-tilt platform, which results in the motion of camera. Both the visual tracker and robot controller form a closed loop for human following task as shown in Figure \ref{overall_framework}.

\subsection{Robot System and Coordinate Definition}

\begin{figure}[htbp]

\begin{minipage}[c]{5cm}
\includegraphics[width=5cm]{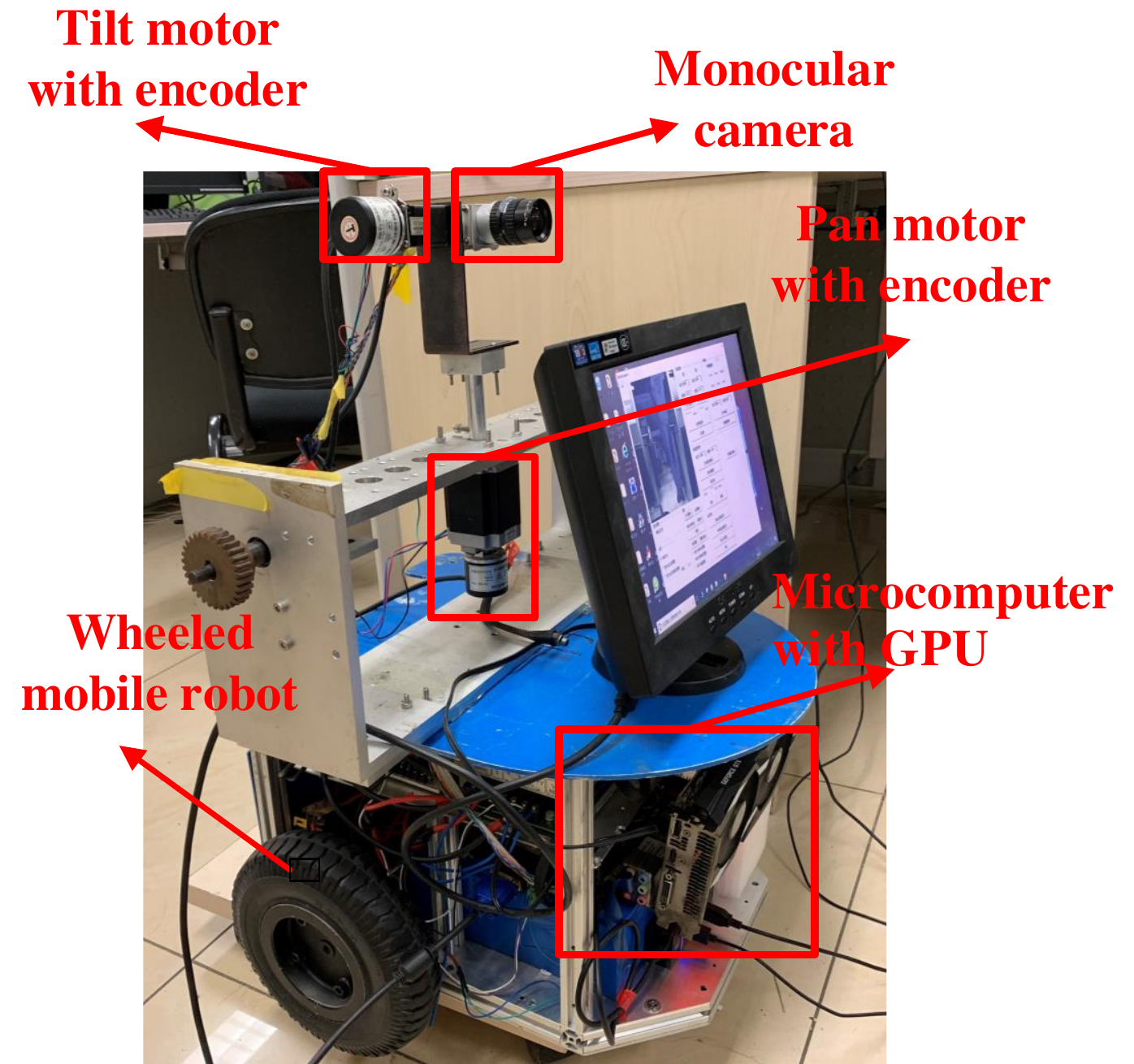}
\end{minipage}%
\begin{minipage}[c]{3.5cm}
\includegraphics[width=3.5cm]{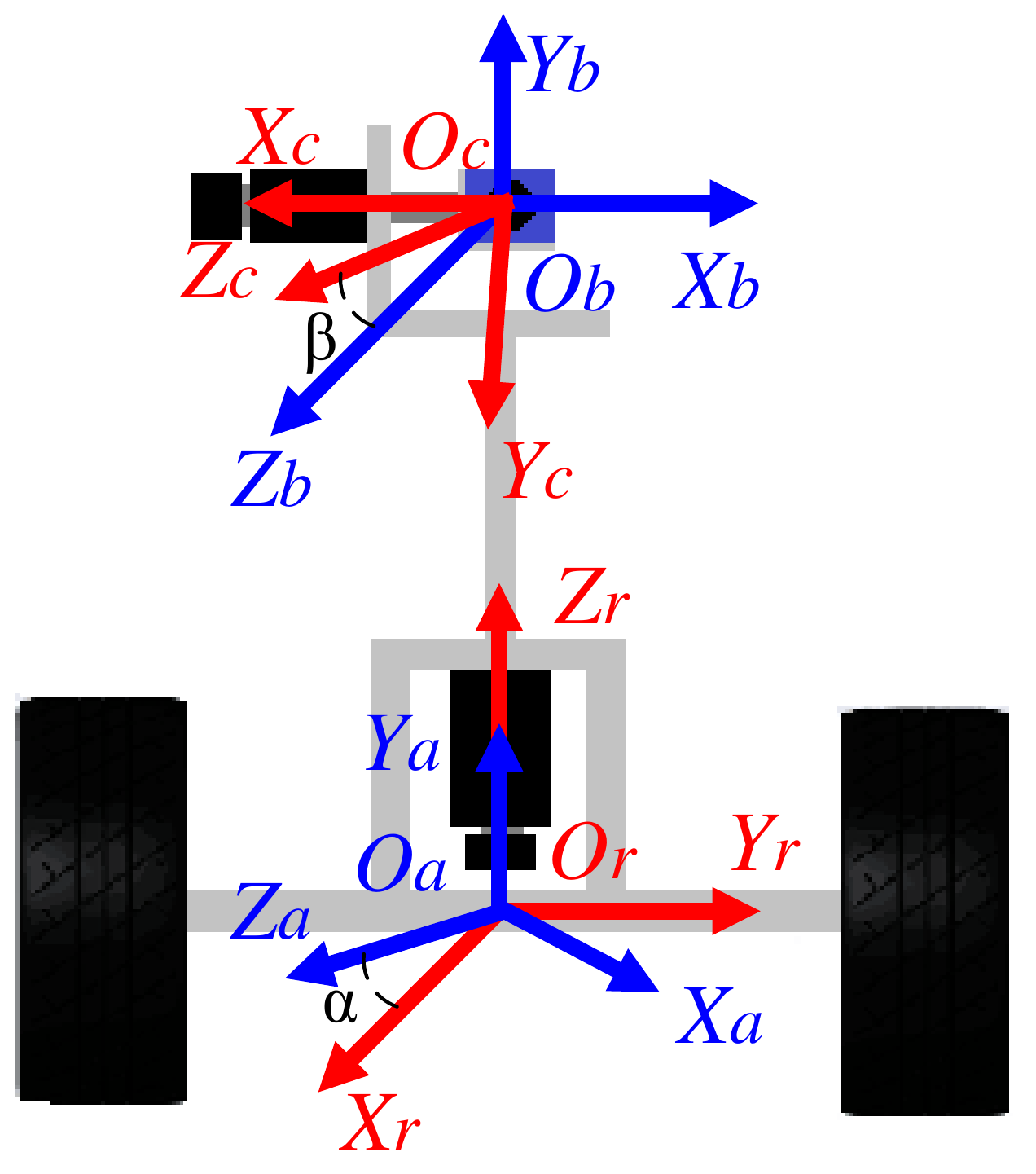}
\end{minipage}%

\caption{\small Left part is the designed robot platform for human following task. Right part is corresponding coordinate systems.}
\label{CoordinateSystem}
\end{figure}

As shown in Figure \ref{CoordinateSystem},
a wheeled mobile robot within monocular pan-tilt camera is adopted in this paper to achieve human following task. The robot system consists of a wheeled mobile platform and a pan-tail camera platform. The wheeled mobile platform are equipped with a microcomputer (including a GTX1050Ti GPU). The pan-tilt camera platform contains pan/tilt motors as well as their corresponding encoders. Compared with conventional human tracking systems that are equipped with fixed depth sensors, our hardware structure has a larger filed-of-view and are less suffered from the perception requirements as well as noises.

Right part of Figure \ref{CoordinateSystem} illustrates the  coordinate systems used in this paper. ${F_r}=(O_r:X_r,Y_r,Z_r)$ represents the mobile robot coordinate system whose origin locates at the midpoint of two wheels axis, and the ${X_r}$-axis is aligned with the forward direction of mobile robot. ${F_a}=(O_a:X_a,Y_a,Z_a)$ and ${F_b}=(O_b:X_b,Y_b,Z_b)$ are defined as pan and tilt coordinate systems respectively. ${F_c}=(O_c:X_c,Y_c,Z_c)$ is the camera coordinate system.
At the beginning, the ${X_a}$-axis is aligned with ${Y_r}$-axis and ${Y_a}$-axis is aligned with ${Z_r}$-axis. The origin of ${F_a}$ is the same with ${F_r}$, and the origin of ${F_c}$ is the same with ${F_b}$.
The direction of ${Z_a}$-axis changes as the pan motor rotates $\alpha$, and the direction of ${Z_c}$-axis changes as the tilt motor rotates $\beta$. All axis of $F_a$ are always aligned with the corresponding axis of $F_b$.

\subsection{Fast Human Detector}

YOLO \cite{YOLO} is utilized as detector in human following task because of its superior speed and accuracy. In this framework, single neural network predicts bounding boxes and class probabilities directly from full images, which regards object detection as a regression problem. Specifically, YOLO divides the full image into $7\times7$ grid
and for each grid cell predicts 2 bounding boxes, confidence for those boxes, and their class probabilities.
For our following task, only \emph{person} class is adopted while other classes are  ignored. We implement Tiny YOLO version on the mobile robot platform where it can perform at 80 FPS. The human bounding box is initialized when the detected position is less than 10 pixels among three consecutive frames.



\begin{figure*}
  \centering
  \includegraphics[width=0.8\linewidth]{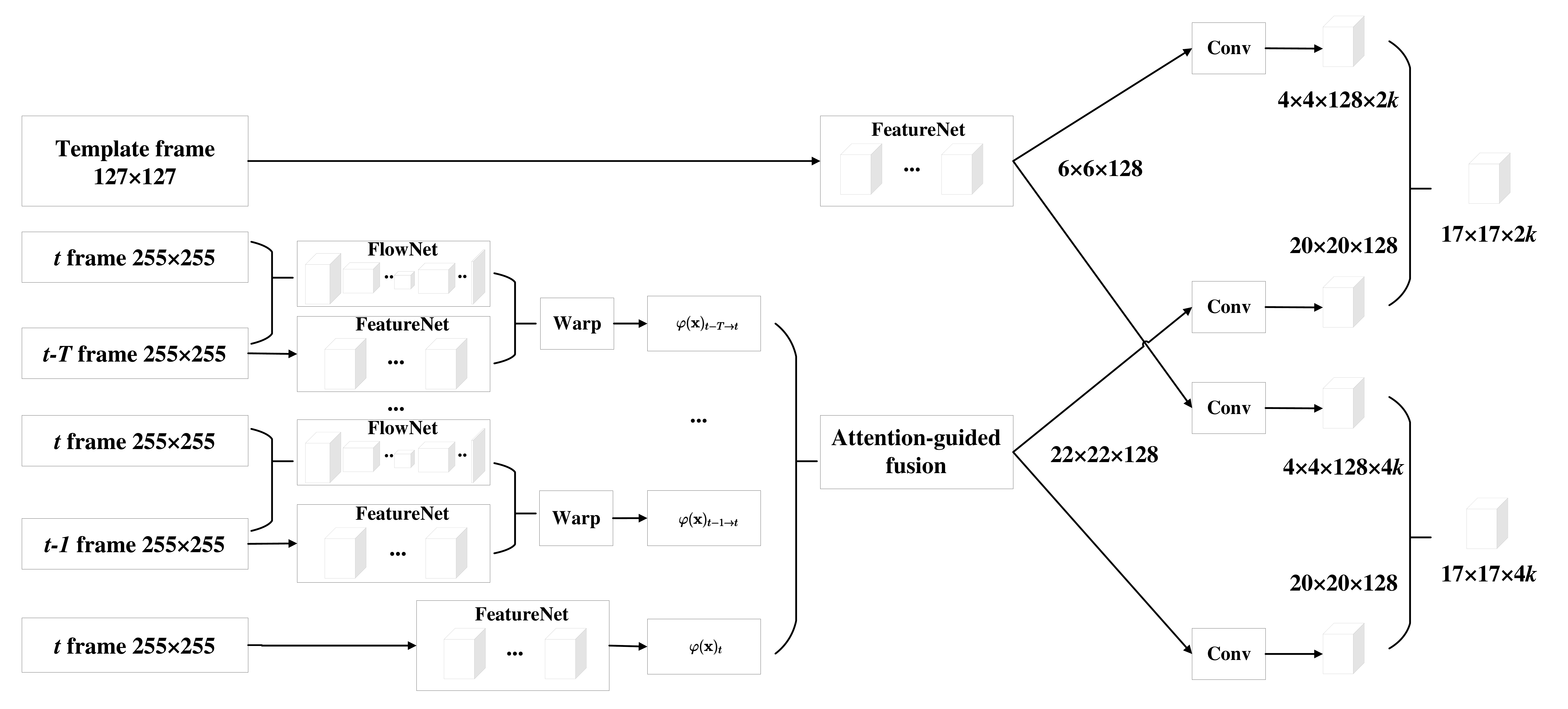}
  \caption{\small The overall framework for FlowTrack++ algorithm. The input size of kernel branch (top) and search branch (bottom) are $127\times127$ and $255\times255$, respectively. The output size of FeatureNet in kernel branch is $6\times6\times128$, which is transformed to a $4\times4\times128\times2k$  kernel and a $4\times4\times128\times4k$ kernel ($k$ is the anchor number in each position) by two $3\times3$ convolution layer. Similarly, the output size of attention-guided fusion in search branch is $22\times22\times128$, which is extended to two $20\times20\times128$ feature maps by $3\times3$ convolution layer. Finally, the feature map and kernels are correlated to produce $17\times17\times2k$ classification map and $17\times17\times4k$ regression map.}
  \label{FlowTrack_framework}
\end{figure*}

\subsection{Real-time and Accurate Visual Tracker}

In this subsection, the overall architecture of proposed \emph{FlowTrack++} algorithm is introduced, which gracefully combines flow aggregation module \cite{FlowTrack} and high-quality Siamese network \cite{SiamRPN, DaSiamRPN}. As shown in Figure~\ref{FlowTrack_framework}, the Siamese network contains a kernel branch and a search branch. In kernel branch, the feature maps of template frame is extracted by FeatureNet. In search branch,
the flow aggregation module contains FeatureNet (feature extraction sub-network), FlowNet \cite{FlowNet}, warping module, attention-guided fusion module.
Appearance features and flow information are extracted by the FeatureNet and FlowNet at first. Then previous frames at predefined intervals is warped to $t$ frame guided by flow information.
Meanwhile, a attention-guided fusion module is designed to weight the warped feature maps.  More details about flow aggregation module can be found in \cite{FlowTrack}.
Finally,  both two branches are fed into subsequent high-quality Siamese network for simultaneous classification and regression \cite{SiamRPN}. All the modules are differentiable and trained end-to-end.

There are always other person and object in the human following scenarios, which may drift the tracking results. Besides, conventional visual tracking algorithms lack consideration for occlusion and out-of-view, which occur frequently in human following.
To address these problems, inspired by \cite{DaSiamRPN}, we adopt hard-negative samples mining strategy and failure recovering strategy.

\paragraph{Hard-negative Samples Mining}
The hard-negative samples contains intra-class pairs and inter-class pairs. In implementations, intra-class pairs are sampled from the different videos that is labelled as \emph{person} (i.e different person from different video). Similarly, inter-class pairs are sampled from the different videos that is labelled different class, such as person and car from different video. All the image pairs are sent to two branches of Figure~\ref{FlowTrack_framework} to train the FlowTrack++ algorithm.
After the hard-negative samples are addressed in training process, the results map of Siamese network becomes high-quality: the high response only appear in the desired target, where the responses of other position (including other human and objects) is suppressed due to the proposed training strategy.

\paragraph{Failure Recovering}
Human following task always encounters occlusion and out-of-view because of the unconstrained environments and drastic camera motions. Conventional trackers lack handling mechanism towards these challenges, which may cause permanent tracking failure. In this paper, a simple yet efficient failure recovering strategy is designed based the high-quality Siamese network output. Specifically, When the failed tracking is indicated (highest score of results map is lower than the threshold), the size of search region is iteratively increased with a constant step size $s$ until the target is re-detected.
This module significantly improves the performance in out-of-view and occlusion challenges. The iterative local-to-global search strategy does not cover the entire images in most cases. This is more efficient than that version of SINT \cite{SINT} which samples over the whole image and adopts time-consuming multi-scale test strategy. The proposed FlowTrack++ algorithm can perform at 40 FPS in human following scenarios.
The detailed process of our failure recovering strategy is described in Algorithm \ref{algorithm1}.

\begin{algorithm}[htbp]
\caption{Algorithm for recovering from tracking failure}
\label{algorithm1}
\begin{algorithmic}[1] %

\REQUIRE thresholds \bm{$Th_{low}$} and \bm{$Th_{high}$} to enter and quit failure cases.\\
\ENSURE target position \bm{$P_t$} and tracking score \bm{$S_t$} during sequences.\\ %
\STATE set \bm{$failure\_state$} = \bm{$False$}
\STATE perform normal tracking in the first
frame, get the position \bm{$P_t$} and the score \bm{$S_t$}.
\REPEAT

\IF{\bm{$S_t \le Th_{low}$}}
\STATE set \bm{$failure\_state$} = \bm{$True$}
\ELSE \IF{\bm{$S_t \ge Th_{high}$}}
       \STATE Set \bm{$failure\_state$} = \bm{$False$}
       \ENDIF
\ENDIF
\IF {\bm{$failure\_state$}}
    \STATE increase the search region by the iterative local-to-global strategy, perform tracking with this larger region, get the position \bm{$P_t$} and the score \bm{$S_t$}.
\ELSE
\STATE perform normal tracking, get the position \bm{$P_t$} and the score \bm{$S_t$}.
\ENDIF
\UNTIL{end of video sequences.}
\end{algorithmic}
\end{algorithm}

\subsection{Unified Controller}




\begin{figure}[htbp]
  \centering
  \includegraphics[width=0.6\linewidth]{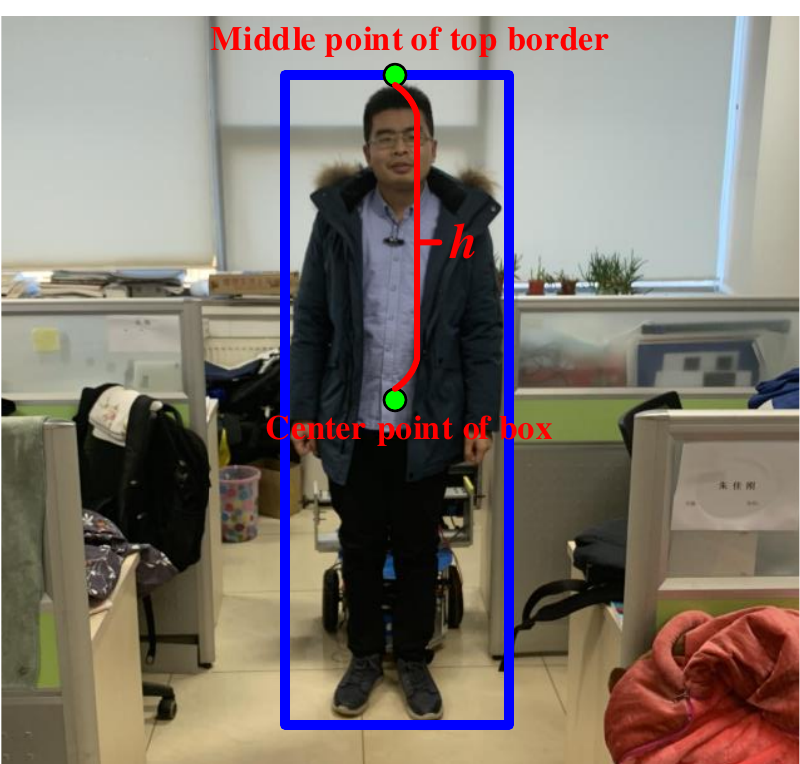}
  \caption{\small The control objective is to stay the human bounding box near the center of view and keep the half height of box ($h$) near a pre-defined constant $H$.}
  \label{human_image}
\end{figure}

In this paper, the controller is designed to stay the human bounding box near the center of view and keep the half height $h$ of the tracking box near a pre-defined constant $H$, which is illustrated in Figure \ref{human_image}. The visual servo formulation is adopted to derive our controller:

\begin{equation}
\mathop {{}^c{P_t}}\limits^.  =  - {}^c{v_c} - {}^c{\omega _c} \times {}^c{P_t}\label{eq2}
\end{equation}
where $ {}^c{P_t} $ is the human target coordinate in camera coordinate system, ${}^c{v_c} $ and ${}^c{\omega _c}$ are the linear and angular velocity of the camera in the camera coordinate, respectively. ${}^c{v_c} $ and ${}^c{\omega _c}$ can be obtained as follows:
\begin{equation}
{}^c{v_c} = {}^c{R_r}{}^r{v_c}={}^c{R_r}\left[ {\begin{array}{*{20}{c}}
{V_r}\\
0\\
0
\end{array}} \right]\label{eq3}
\end{equation}

\begin{equation}
{}^c{\omega _c} = {}^c{R_r}\left[ {\begin{array}{*{20}{c}}
0\\
0\\
{{\omega _r} + {\omega _\alpha }}
\end{array}} \right] + \left[ {\begin{array}{*{20}{c}}
{{\omega _\beta }}\\
0\\
0
\end{array}} \right]\label{eq4}
\end{equation}
where $^c{R_r}$ denotes the rotation matrix from ${F_r}$ to ${F_c}$, $V_{r}$ is the velocity of the mobile robot, ${\omega _\alpha }$ and ${\omega _\beta }$ denote the the pan and tilt angular velocity of the camera respectively, and ${\omega _r}$ represents the robot angular velocity.


Substituting (\ref{eq3}) and (\ref{eq4}) into (\ref{eq2}), equation (\ref{eq2}) can be rewritten as follows:

\begin{equation}
\!\!\!\!\!\!\left[ \!\!\!{\begin{array}{*{20}{c}}
{\mathop {{}^cx}\limits^. }\\
{\mathop {{}^cy}\limits^. }\\
{\mathop {{}^cz}\limits^. }{}
\end{array}}\!\! \right] \!\!= \!\!\left[ \!\!\!{\begin{array}{*{20}{c}}
{ -{V_r}\sin \alpha  + ({\omega _\alpha } + {\omega _r})\cos \beta {}^cz - ({\omega _\alpha } + {\omega _r})\sin \beta {}^cy}\\
{ {V_r}\cos \alpha \sin \beta  + ({\omega _\alpha } + {\omega _r})\sin \beta {}^cx - {\omega _\beta}{}^cz}\\
{ - {V_r}\cos \alpha cos\beta  + {\omega _\beta}{}^cy - ({\omega _\alpha } + {\omega _r})\cos \beta {}^cx}
\end{array}} \!\!\!\right]\!\!\!\label{eq5}
\end{equation}
 The control objective is to stay the human bounding box near the center of view and keep the half height $h$ of box near a pre-defined constant $H$. To this end, the three image errors in pixel level are defined as:
\begin{equation}
\left\{ \begin{array}{l}
{e_u} = u - {u_0}\\
\\
{e_v} = v - {v_0}\\
\\
{e_{v2}} = v - {v_2}-H
\end{array} \right.\label{eq6}
\end{equation}
 where ${({u},{v})}$ is the center point of human bounding box, ${({u},{v_2})}$ is the middle point of top border, ${({u_0},{v_0})}$ is the center point of captured image, $H$ is a pre-defined constant, which is the half of the desired height of tracking box.
The relationship between coordinate systems ${F_b}$ and ${F_c}$ is:
\begin{equation}
\left[ {\begin{array}{*{20}{c}}
{{}^b x}\\
{{}^b y}\\
{{}^b z}\\
1
\end{array}} \right] = \left[ {\begin{array}{*{20}{c}}
1&0&0&0\\
0&{\cos \beta }&{ - \sin \beta }&0\\
0&{\sin \beta }&{\cos \beta }&0\\
0&0&0&1
\end{array}} \right]{\rm{ }}\left[ {\begin{array}{*{20}{c}}
{{}^cx}\\
{{}^cy}\\
{{}^cz}\\
1
\end{array}} \right]\label{eq8}
\end{equation}

 Since human and mobile robot move on the flat ground, we can assume that the hight of human in coordinate system ${F_b}$, i.e
 $^{b}y$ remains constant.
 According to equation (\ref{eq6}), (\ref{eq8}) and pin-hole camera model (${\alpha _x}$, ${\alpha _y}$, ${u_0}, {v_0}$ are camera intrinsic parameters, $(^cx, ^cy, ^cz)$ represents a point in camera coordinate system and ${({u},{v})}$ is its corresponding image coordinate), the relationship between $^cz$ and $^b y$ can be obtained:

\begin{equation}
{{}^cz} = \frac{{{\alpha_y \cdot {^by}}}}{{{e_v}\cos \beta  - {\alpha _y}\sin \beta }}\label{eq10}
\end{equation}
 \newcounter{mytempeqncnt}
\begin{figure*}[ht]
\footnotesize
\setcounter{mytempeqncnt}{\value{equation}}
\setcounter{equation}{12}
\begin{equation}
\tiny
\mathop {V_r} =- \frac{(BC - AD)({K_2}{e_{v}}-{K_3}{e_{v2}})  +(AF-BE){K_2}{e_{v}} - (CF-DE){K_1}{e_{u}}}
 {(BC - AD){\Omega _3}{\lambda_2}+(AF-BE){\Omega _2}{\lambda_1}-(CF-DE){\Omega _1}{\lambda_1}}
 \label{c1}
\end{equation}
\begin{equation}
\tiny
\!\!\!\!\!\!\!\!\!\!\!\!\!\!\!\!\!\!\!\!\!\!\!\!\!\!\!\mathop {\omega _\alpha } \!=\!\frac{(D{K_1}{e_{u}}\!\!-\!\! B{K_2}{e_{v}}\!\!-\!\! BC{\omega _r}\!\!+\!\!AD{\omega _r}){\Omega _3}{\lambda_2}
\!\!+\!\!(BE{\omega _r}\!\!-\!\! AF{\omega _r}\!\!-\!\! B{K_3}{e_{v2}}\!\!+\!\!B{K_2}{e_{v}}\!\!-\!\! F{K_1}{e_{u}}) {\Omega _2}{\lambda_1}
\!\!+\!\!(CF{\omega _r}\!\!-\!\! DE{\omega _r}\!\!+\!\! D{K_3}{e_{v2}}\!\!-\!\! D{K_2}{e_{v}}\!\!+\! \!F{K_2}{e_{v}}) {\Omega _1}{\lambda_1}}
 {(BC \!\!-\!\! AD){\Omega _3}{\lambda_2}\!\!+\!\!(AF\!-\!BE){\Omega _2}{\lambda_1}\!\!-\!\!(CF\!\!-\!\!DE){\Omega _1}{\lambda_1}}
 \label{c2}
\end{equation}
\begin{equation}
\tiny
\!\!\!\!\!\!\mathop {\omega _\beta } = \frac{(C{K_3}{e_{v2}}-C{K_2}{e_{v}}+E{K_2}{e_{v}}){\Omega _1}{\lambda_1}- (A{K_2}{e_{v}}{\lambda_2}-C{K_1}{e_{u}}){\Omega_3}{\lambda_2}-(A{K_2}{e_{v}}+A{K_3}{e_{v2}}+E{K_1}{e_{u}}){\Omega_2}{\lambda_1}}
 {(BC - AD){\Omega _3}{\lambda_2}+(AF-BE){\Omega _2}{\lambda_1}-(CF-DE){\Omega _1}{\lambda_1}}
 \label{c3}
\end{equation}
\begin{equation}
\tiny
{\omega _{\rm{r}}} = \left\{ \begin{array}{l}
0,{\rm{       }} - \frac{\pi }{6}{\rm{  <  }}\alpha  < \frac{\pi }{6}\\
0.1\alpha ,{\rm{  }}\alpha  <  - \frac{\pi }{6}{\rm{ \ or \ }}\alpha  > \frac{\pi }{6}
\end{array} \right.
\label{c4}
\end{equation}
\setcounter{equation}{\value{mytempeqncnt}}
\hrulefill
\vspace*{4pt}
\end{figure*}

\normalsize

Differentiating equation (\ref{eq6}) to time on both sides and take (\ref{eq10}) into consideration, we can obtain equation (\ref{eq11}).
\begin{equation}
\left\{ \begin{array}{l}
\mathop {{e_u}}\limits^.  = \lambda_1{V_r}{\Omega _1} + A({\omega _\alpha } + {\omega _r}) + B{\omega _\beta }\\
\\
\mathop {{e_v}}\limits^.  = \lambda_1{V_r}{\Omega _2} + C({\omega _\alpha } + {\omega _r}) + D{\omega _\beta }\\
\\
\mathop {{e_{v2}}}\limits^.  = \lambda_1{V_r}{\Omega _2} + C({\omega _\alpha } + {\omega _r})
+ D{\omega _\beta } - \lambda_2{V_r}{\Omega _3}
\\ - E({\omega _\alpha } + {\omega _r}) -  F\omega_\beta
\end{array} \right.
\label{eq11}
\end{equation}
where $\lambda_1=\frac{1}{{{}^b y_1}}$ and $\lambda_2=\frac{1}{{{}^b y_2}}$, $^by_1$ is the y-coordinate of the center point of human body, and
$^by_2$ is the y-coordinate of the middle point of top border as shown in Figure \ref{human_image}. They are both donated in coordinate system ${F_b}$. The meaning of other symbols are shown in equation (\ref{eq1000}).

\begin{equation}
\left\{ \begin{array}{l}
{\Omega _1} = \frac{{1}}{{{\alpha _y}}}({\alpha _x}\sin \alpha - {e_v}\cos \alpha cos\beta   )({e_v}cos\beta  + {\alpha _y}\sin \beta ) \\
\\
{\Omega _2} = \frac{{1}}{{{\alpha _y}}}({e_v}\cos \alpha cos\beta  + {\alpha _y}cos\alpha \sin \beta )\cdot\\(-{e_v}cos\beta  - {\alpha _y}\sin \beta )\\
\\
{\Omega _3} = \frac{{1}}{{{\alpha _y}}}((v_2-v_0)\cos \alpha cos\beta  + {\alpha _y}cos\alpha \sin \beta )\\(-(v_2-v_0)cos\beta  - {\alpha _y}\sin \beta )\\
\\
 A ={\rm{ }}\frac{1}{{{\alpha _x}{\alpha _y}}}({{\alpha _x}^2{\alpha _y}\cos \beta  - {\alpha _x}^2\sin \beta {e_v} + {e_u}^2{\alpha _y}\cos \beta })\\
\\
B =-\frac{1}{{{\alpha _y}}}({{e_u}{e_v}}) \\
 \\
 C =\frac{1}{{{\alpha _x}}}({{\alpha _y}\sin \beta {e_u} + {e_u}{e_v}\cos \beta })\\
\\  D= - \frac{1}{{{\alpha _y}}}({\alpha _y^2 + e_v^2})\\
\\ E =\frac{1}{{{\alpha _x}}}({{\alpha _y}\sin \beta (u_2-u_0) - (u_2-u_0)(v_2-v_0)\cos \beta })\\
\\ F= - \frac{1}{{{\alpha _y}}}({\alpha _y^2 + e_{v2}^2})
\end{array} \right.
\label{eq1000}
\end{equation}

To make errors converge to zero, the following equations should be satisfied:

\begin{equation}
\left\{ \begin{array}{l}
\mathop {{e_u}}\limits^.  = -K_1e_u\\
\\
\mathop {{e_v}}\limits^.  = -K_2e_v\\
\\
\mathop {{e_{v2}}}\limits^.  = -K_3e_{v_2}
\end{array} \right.\label{eq100}
\end{equation}
where $K_1$, $K_2$ and $K_3$ are positive gains respectively. Substituting (\ref{eq11}) and (\ref{eq1000}) into (\ref{eq100}), robot linear velocity, pan and tilt angular velocity of the camera can be obtained by
equations (\ref{c1}), (\ref{c2}), and (\ref{c3}) respectively. Besides, the strategy to control the robot angular velocity $\omega_r$ is adopted as equation (\ref{c4}).

\normalsize
\section{Experiments}

\subsection{Simulation Results of Control Law}

\begin{figure}[!tp]
 \centering
\begin{minipage}[c]{3cm}
\includegraphics[width=3cm]{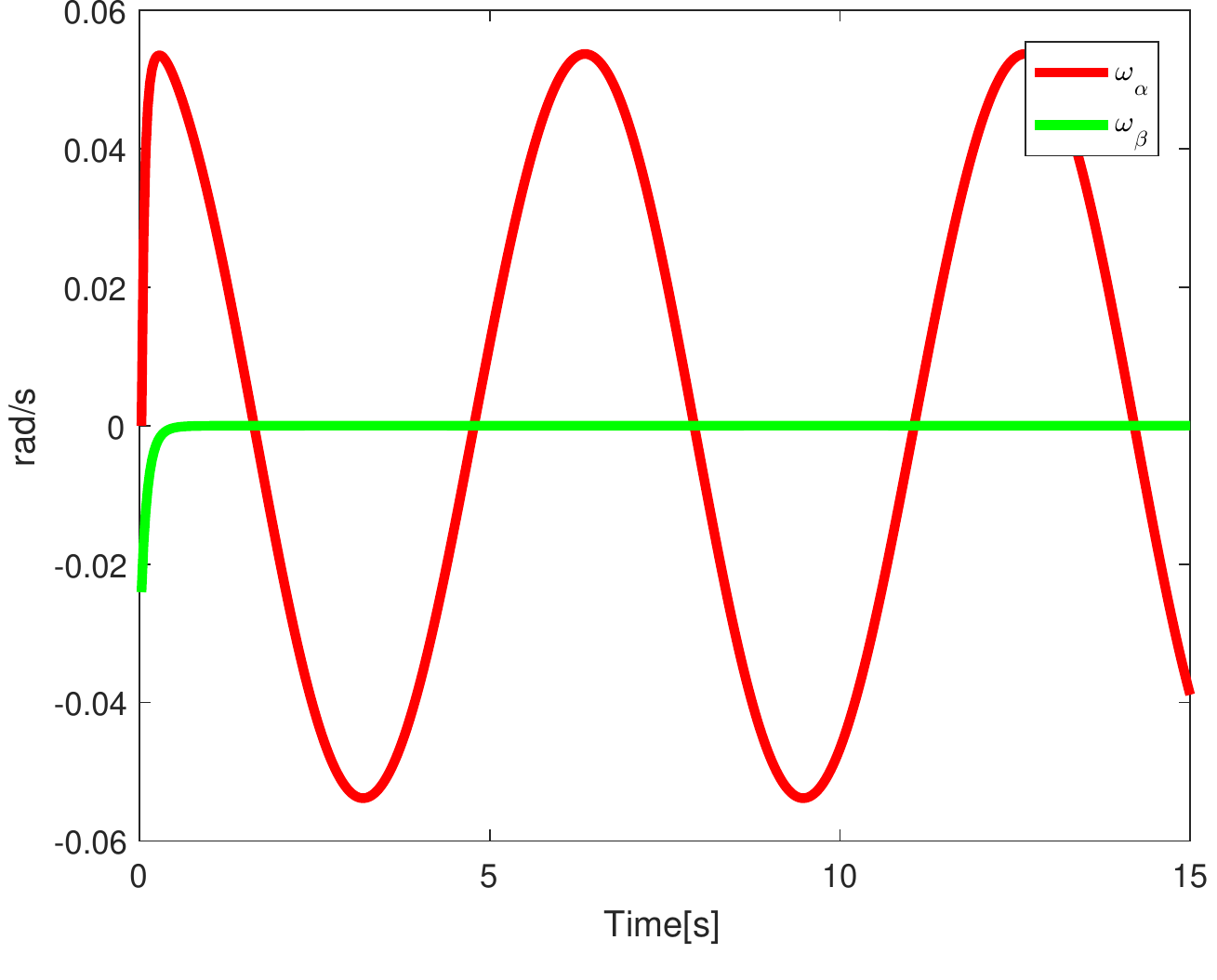}
\label{sim_1}
\end{minipage}%
\begin{minipage}[c]{3cm}
\includegraphics[width=3cm]{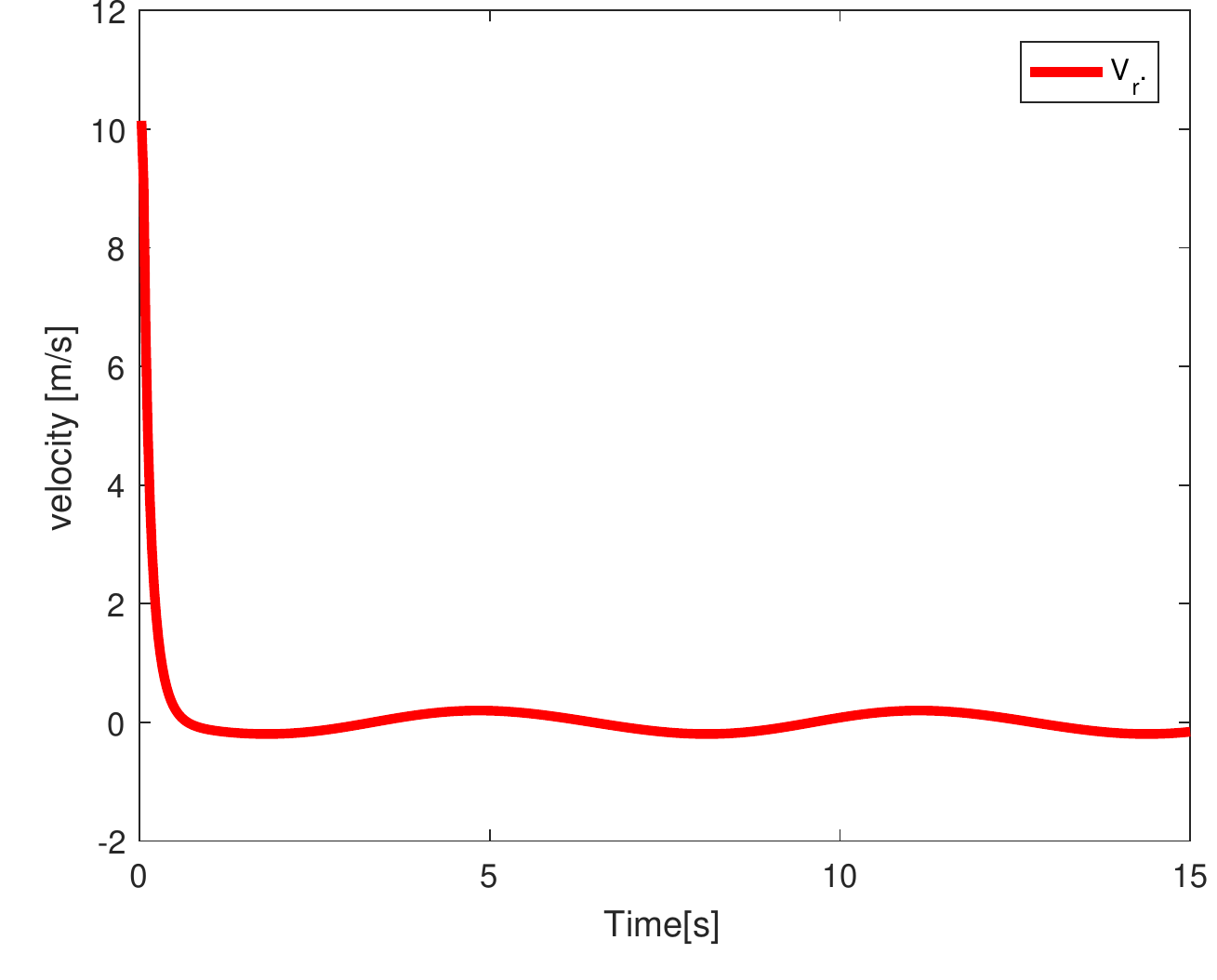}
\label{sim_2}
\end{minipage}%

\begin{minipage}[c]{3cm}
\includegraphics[width=3cm]{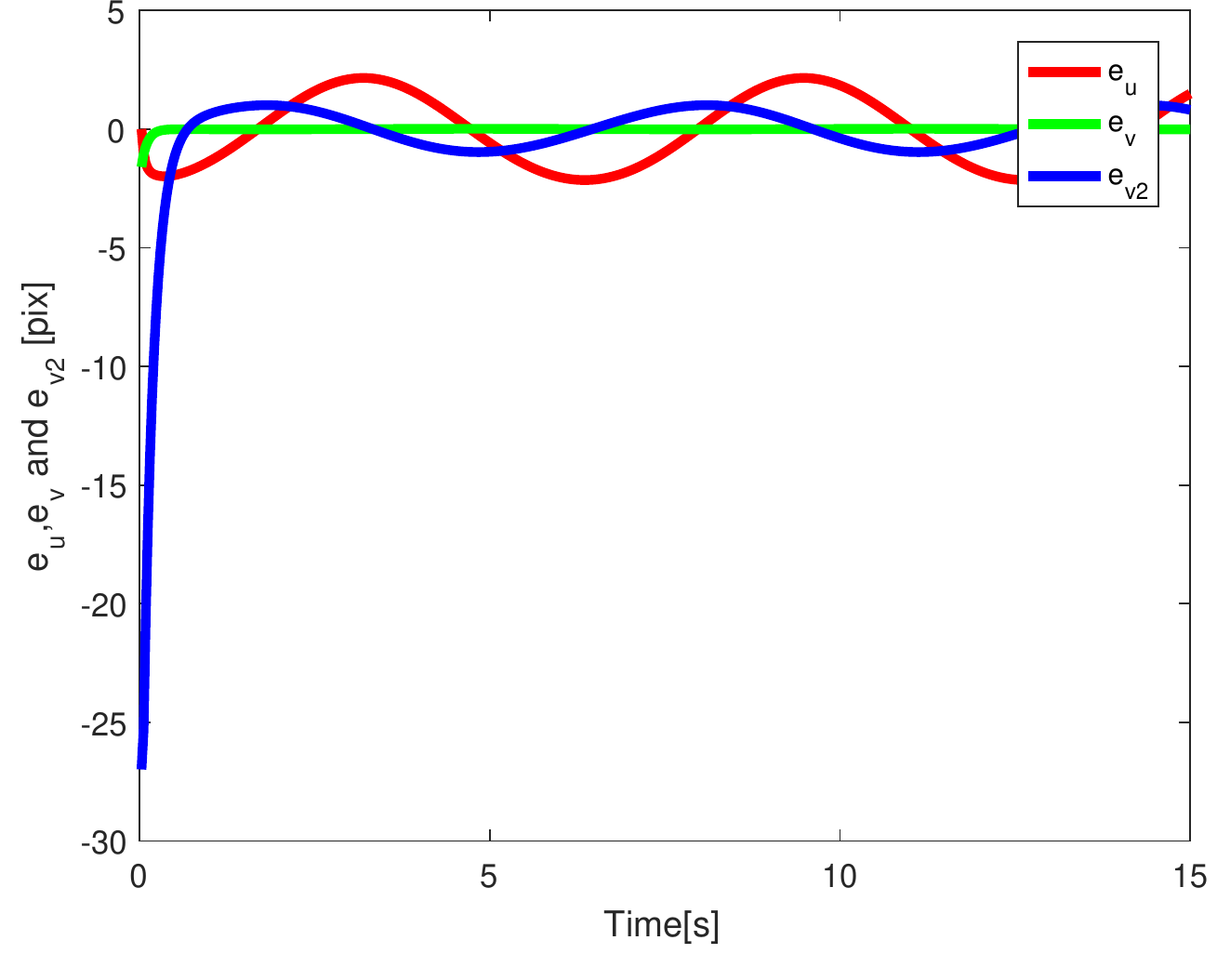}
\label{sim_3}
\end{minipage}%
\begin{minipage}[c]{3cm}
\includegraphics[width=3cm]{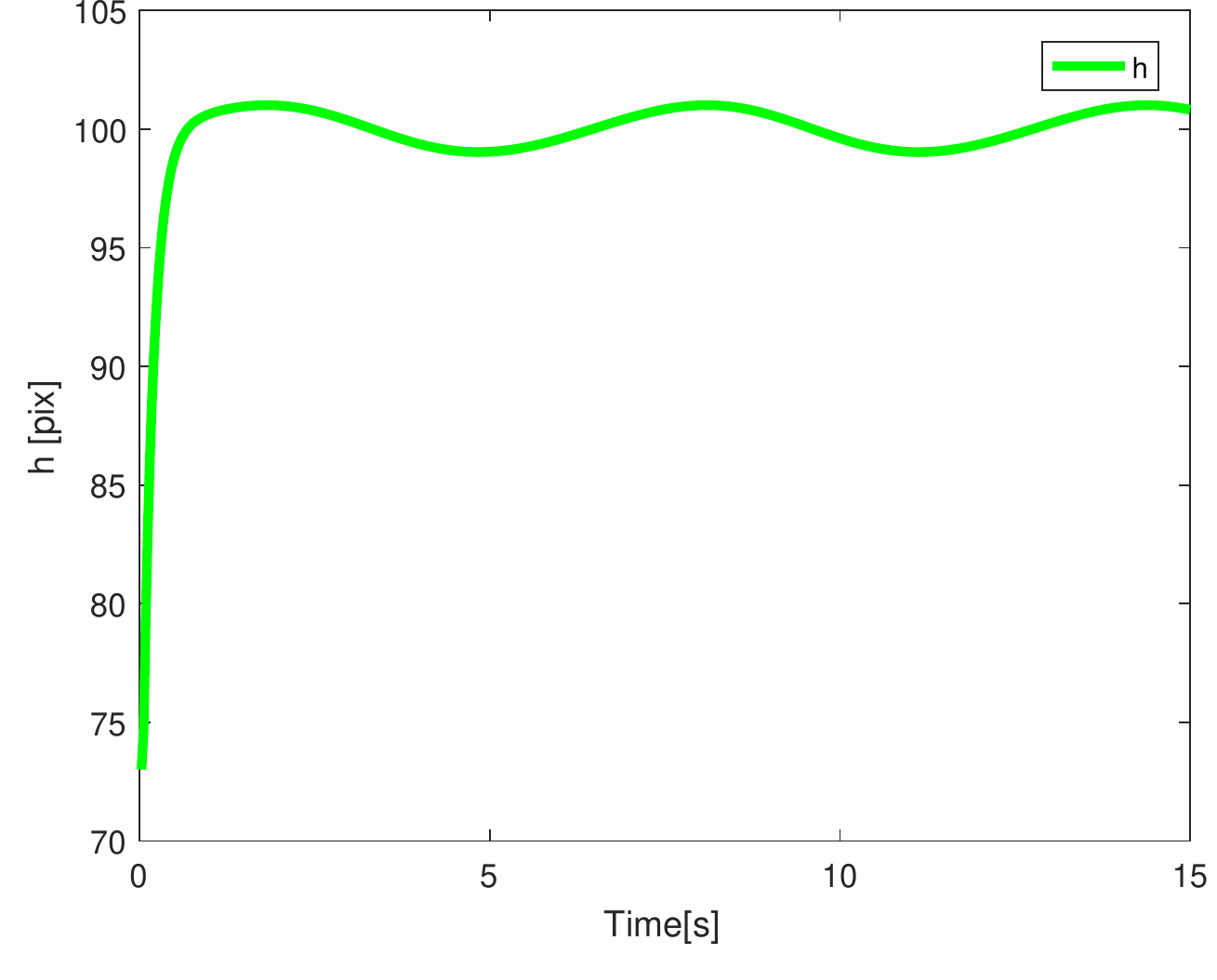}
\label{sim_4}
\end{minipage}%
 \caption{\small The simulation result when the target moves along a circle, including the curves of $\omega_\alpha$ and $\omega_\beta$, the curve of $V_r$, the curves of $e_u$, $e_v$ and $e_{v2}$, the curve of $h$.}
 \label{sim2}
\end{figure}

In this section, simulations are performed to verify the effectiveness of proposed controller. A rectangle is utilized to simulate the tracked human, where the control objective is to stay the human bounding
box near the center of view and keep the half height of box near a pre-defined constant $H$. In our simulation, the human moves along a circle in world coordinate system. The center of the circle is at $(0.5, 0.5)$, and it is denoted as the following equation:

\setcounter{equation}{16}
\begin{equation}
\left\{ \begin{array}{l}
 x_t = 0.5- 0.4\cos(t)\\
\\
 y_t = 0.5 - 0.4\sin(t)

\end{array} \right.\label{sim_2}
\end{equation}
The robot starts moving from $(0, 0)$ in the world coordinate system and $H$ is set to 100. As shown in Figure \ref{sim2} the proposed controller results in a smooth trajectory. Because the object is moving along an circle, the $\omega_\alpha$ changes in a sinusoidal wave. The height of the human in image changes little, so the $\omega_\beta$ changes a little. As the human moves, in order to adjust the size of rectangle in the image, the mobile robot moves forward and backward, and the value of $V_r$ fluctuates slightly around 0. Due to motion of the human, $e_u$, $e_{v}$ and $e_{v2}$ are always changing. Besides, $h$ can stay around 100 according to the designed controller. From above results, one can find that the control law is effective.

\subsection{Human Following Results}
In this section, the effectiveness of human following system is verified by indoor and outdoor experiments on real-world robot platform. The indoor experiment is conducted in our laboratory and the outdoor experiment is  performed outside the building in the Institute of Automation, Chinese Academy of Sciences. The height of the mounted camera is 0.7m, and the height of the person is about 1.8m. So the $\lambda_1$ and $\lambda_2$ in equation (\ref{eq11}) are 5 and 0.91, respectively. We compared three different tracking algorithms on our designed controller, including proposed FlowTrack++, ECO \cite{ECO} and GOTURN \cite{GOTURN}.

\begin{figure}
\centering
\subfigure[]{
\includegraphics[width=0.8\linewidth]{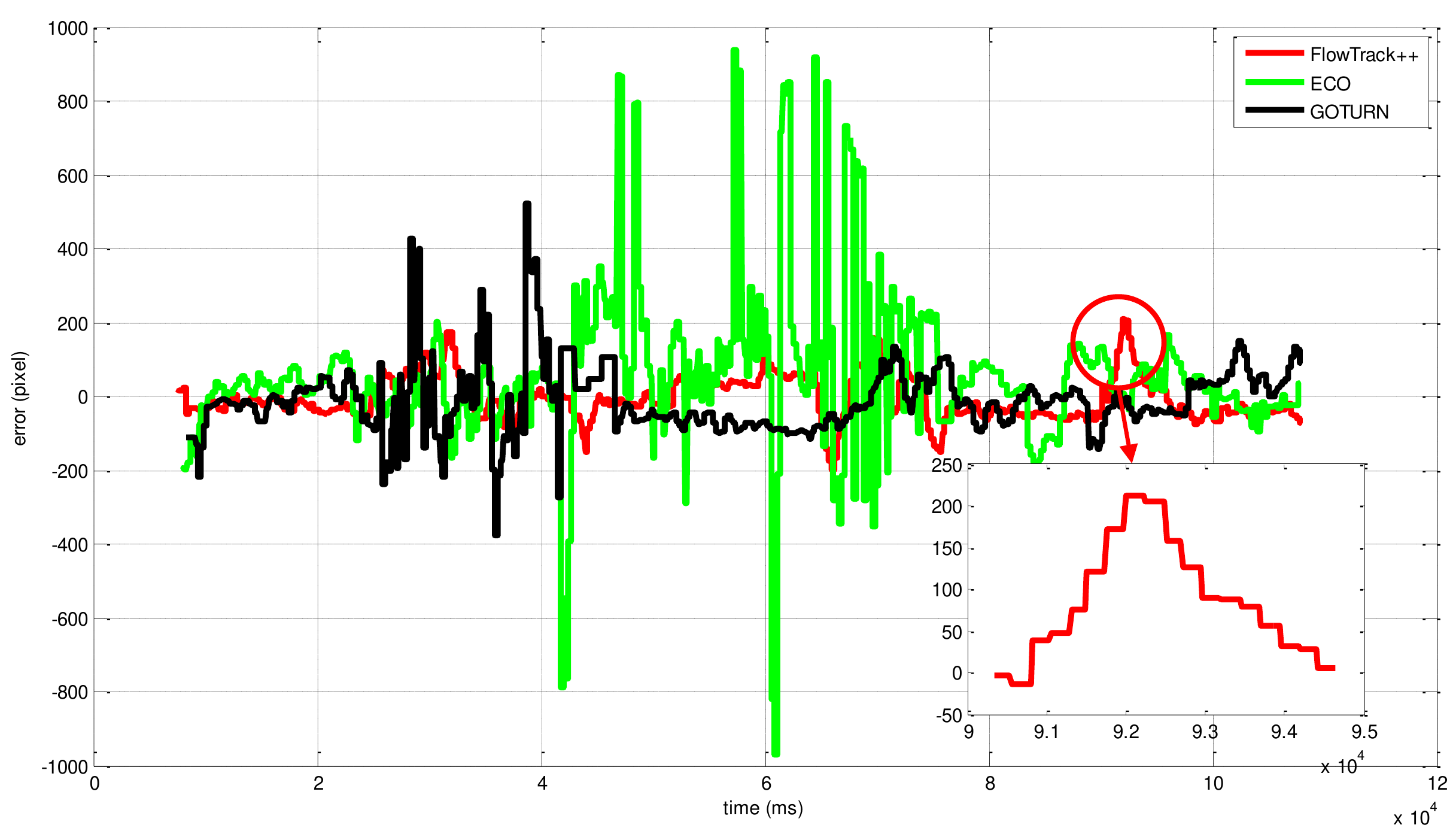}
\label{indoor1}
}
\subfigure[]{
\includegraphics[width=0.8\linewidth]{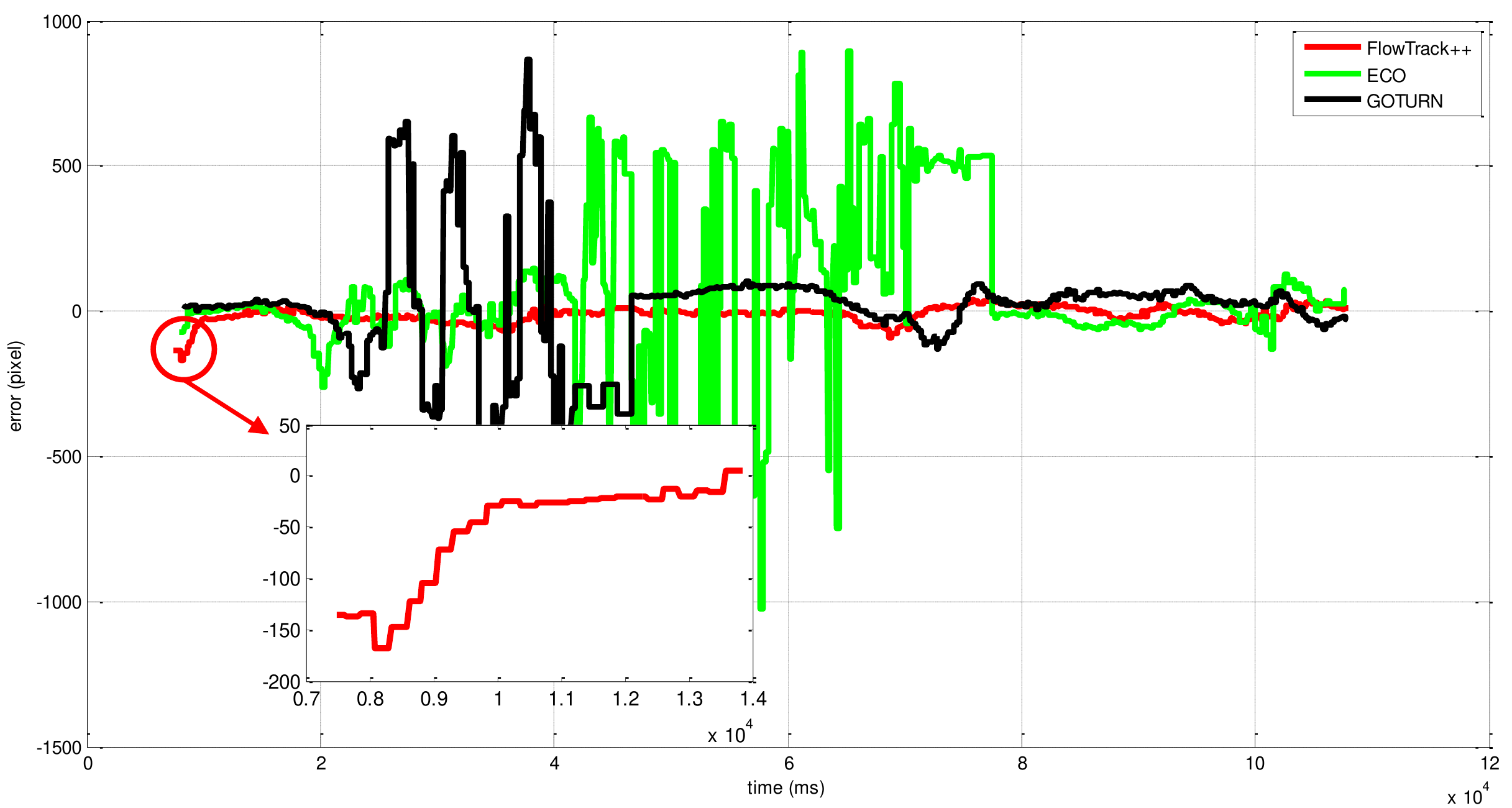}
\label{indoor2}
}
\subfigure[]{
\includegraphics[width=0.8\linewidth]{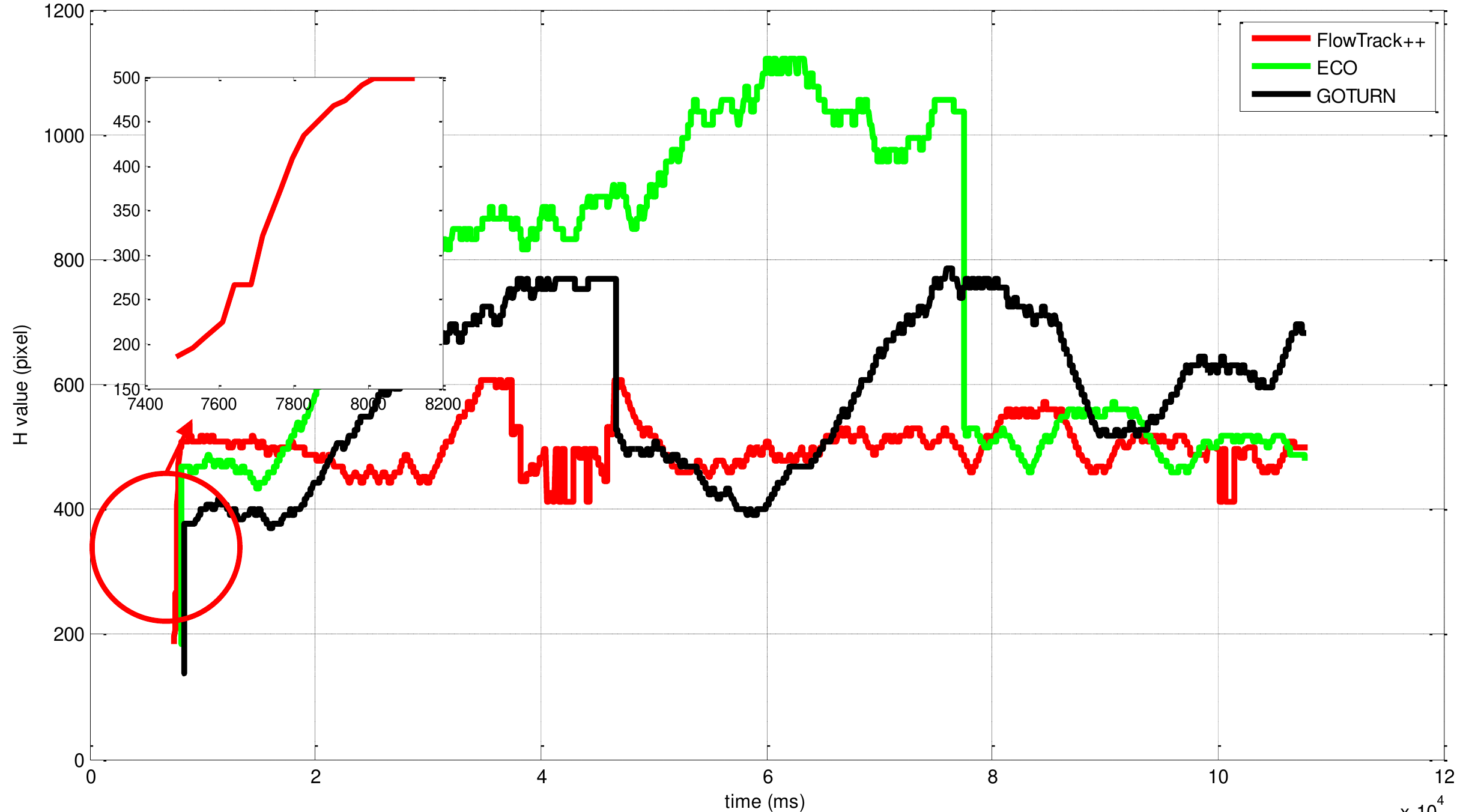}
\label{indoor3}
}
\caption{\small The results of indoor experiments. (a) Curves of $e_u$. (b) Curves of $e_v$. (c) Curves of $h$.}
\label{indoor}
\end{figure}

\begin{figure}
\centering
\subfigure[]{
\includegraphics[width=0.8\linewidth]{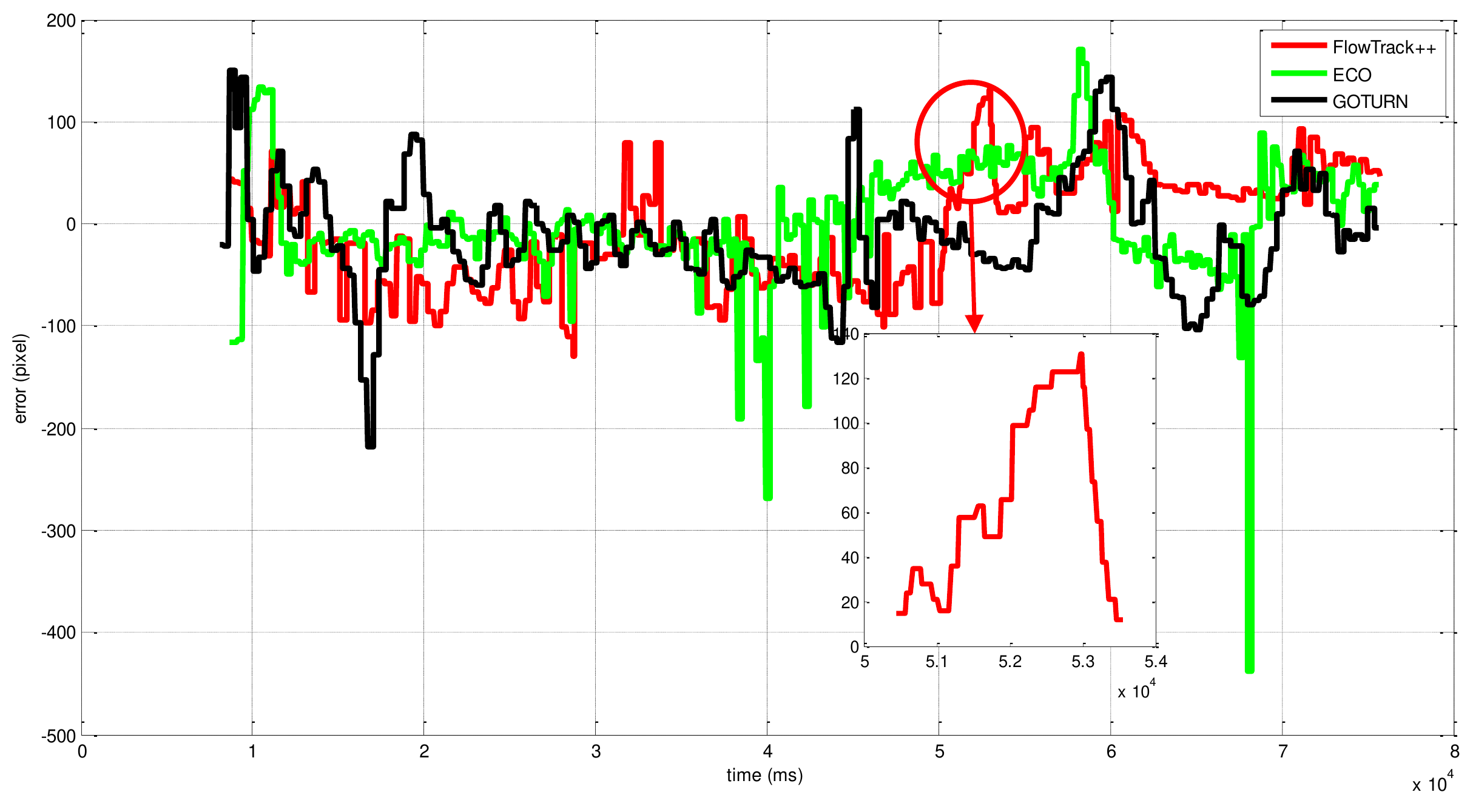}
\label{outdoor1}
}
\subfigure[]{
\includegraphics[width=0.8\linewidth]{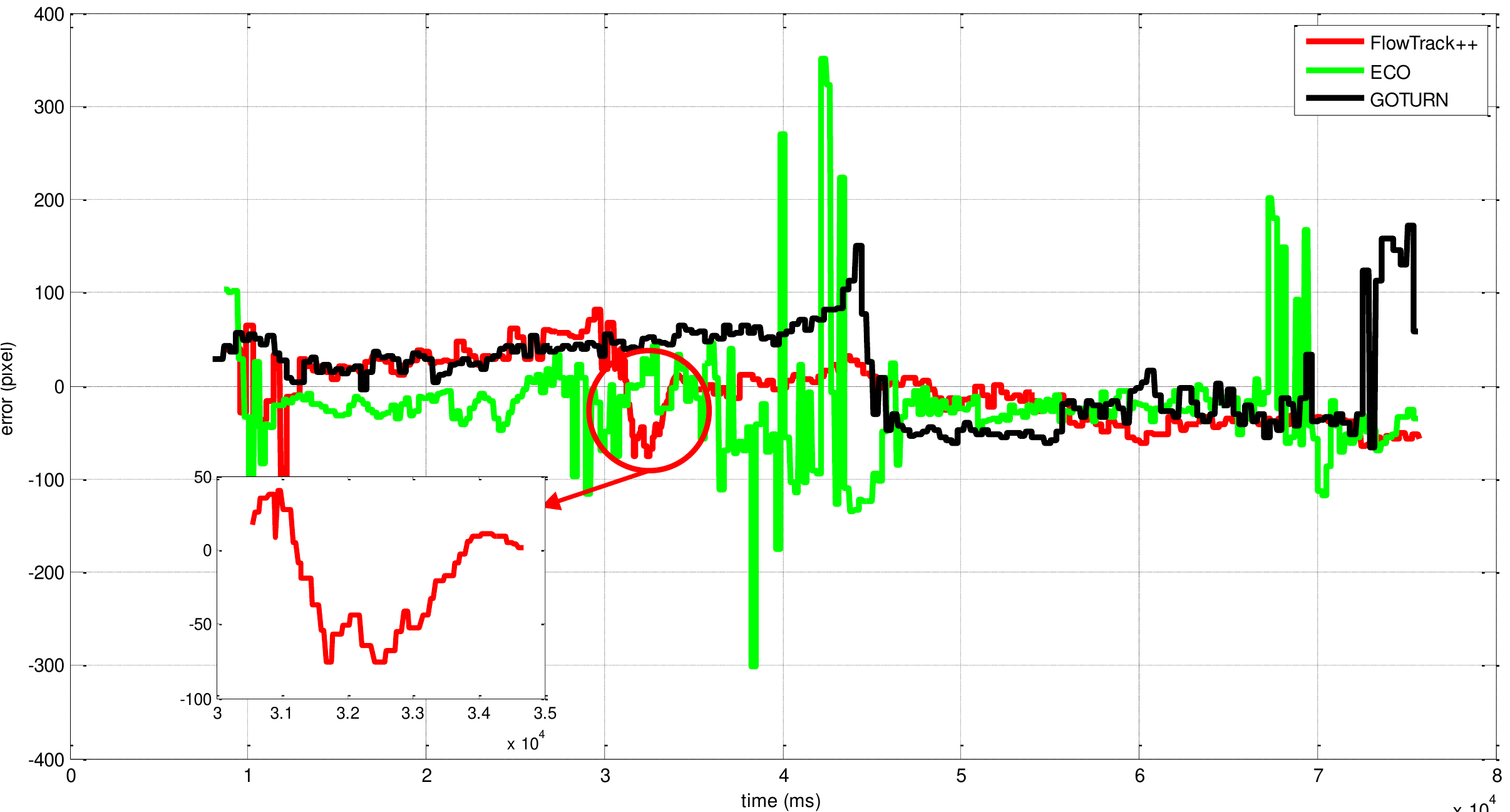}
\label{outdoor2}
}
\subfigure[]{
\includegraphics[width=0.8\linewidth]{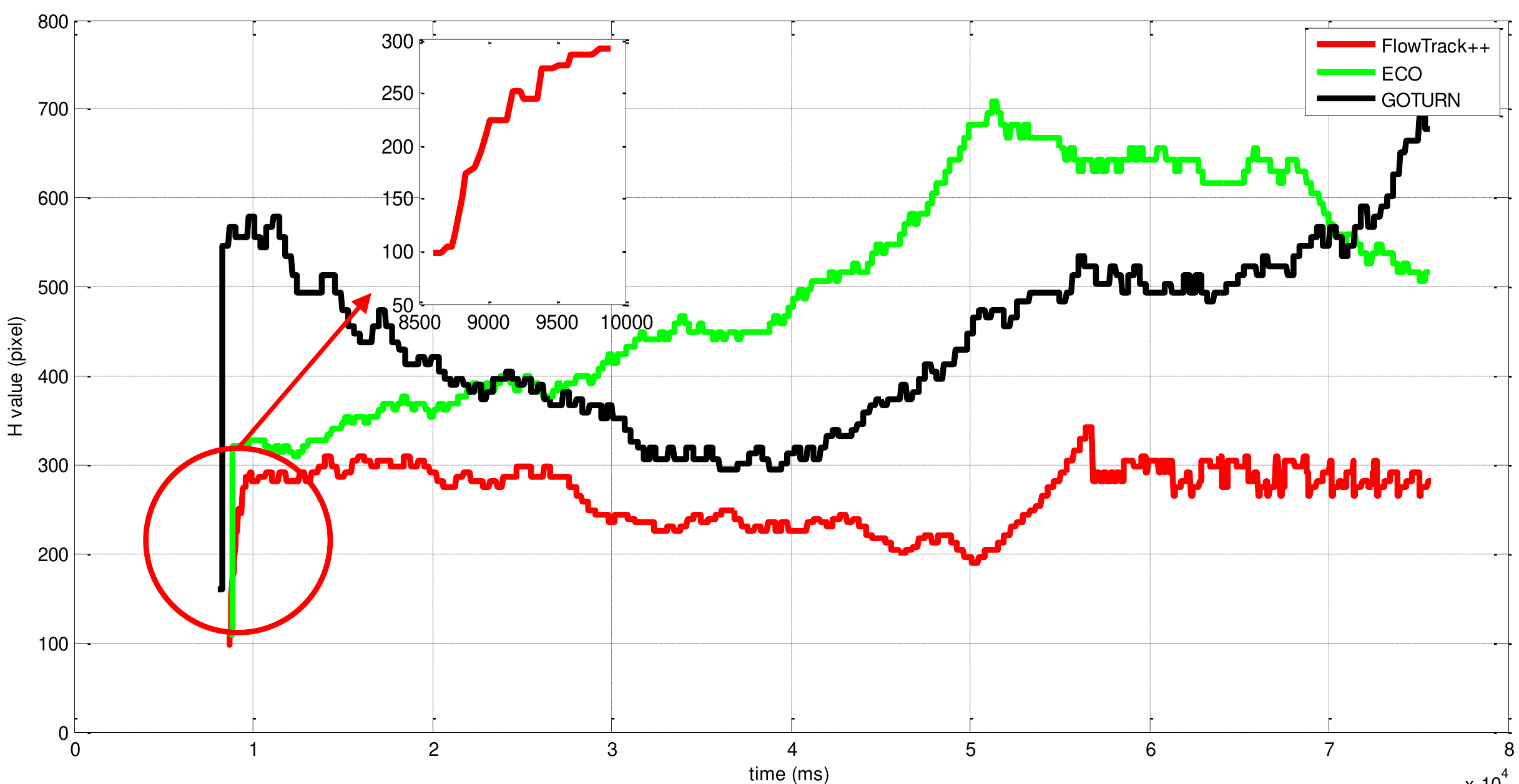}
\label{outdoor3}
}
\caption{\small The results of outdoor experiments. (a) Curves of $e_u$. (b) Curves of $e_v$. (c) Curves of $h$.}
\label{outdoor}
\end{figure}

In indoors experiments, $H$ is set as 500 pixel. Figure \ref{indoor} shows the influence of different tracking algorithm on the controller. When adopting proposed FlowTrack++, the $e_u$ increases to 211 pixels due to the relative motion between humans and robots while $t=92s$. Then when $t=95s$, $e_u$ decreases to 0 quickly. Besides, the tracked human hight $h$ can stay around 500 pixels while other trackers fluctuate violently.
From those results, it can be known that the tracking algorithms of \emph{FlowTrack++} have better performance on human following task.

\begin{figure}
\centering
\subfigure[]{
\includegraphics[width=0.8\linewidth]{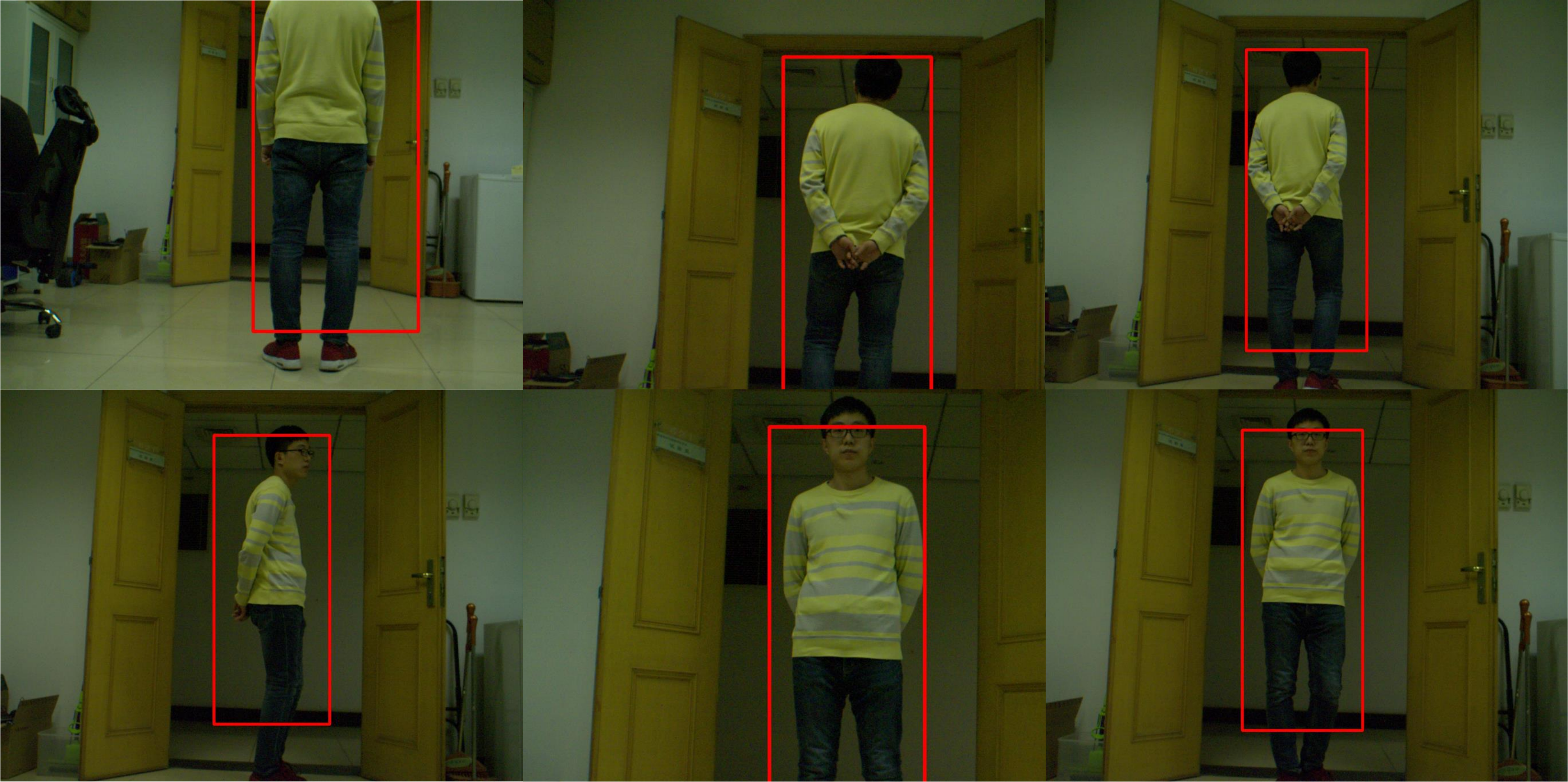}
\label{following_pose}
}
\subfigure[]{
\includegraphics[width=0.8\linewidth]{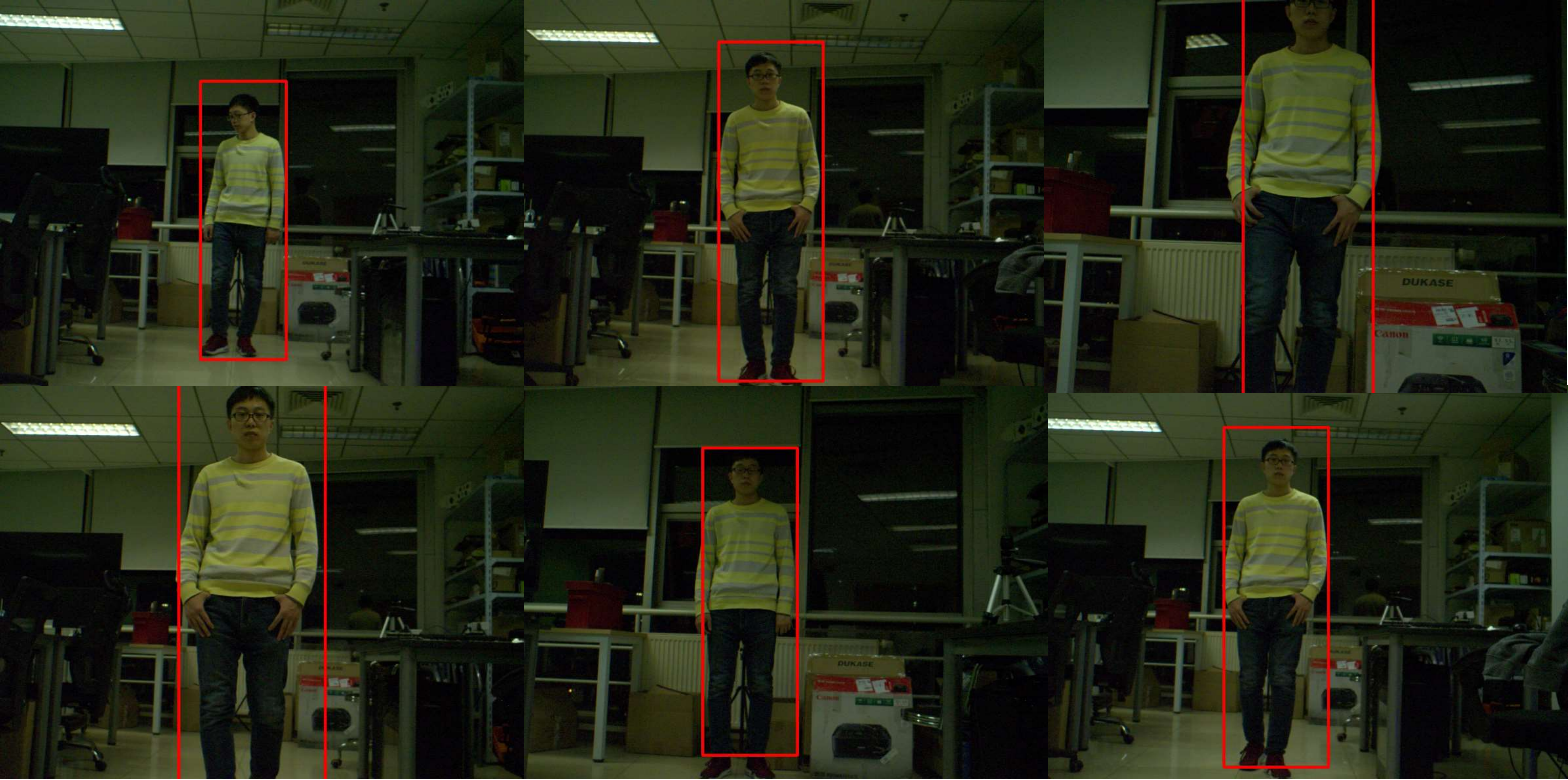}
\label{following_scale}
}
\caption{\small Human tracking results in indoor experiments. (a) Pose changes challenge. (b) Scale challenge.}
\label{indoor_results}
\end{figure}

In the outdoor experiments as shown in Figure \ref{outdoor}, $H$ is set as 300 pixel. By \emph{FlowTrack++} tracking algorithm, $e_u$ and $e_v$ can be kept around 0, and the $h$ fluctuates stably around 300 pixels. Compared with FlowTrack++ algorithm, both ECO and GOTURN trackers result in larger tracking errors and unstable following results.

\begin{figure}
\centering
\subfigure[]{
\includegraphics[width=0.8\linewidth]{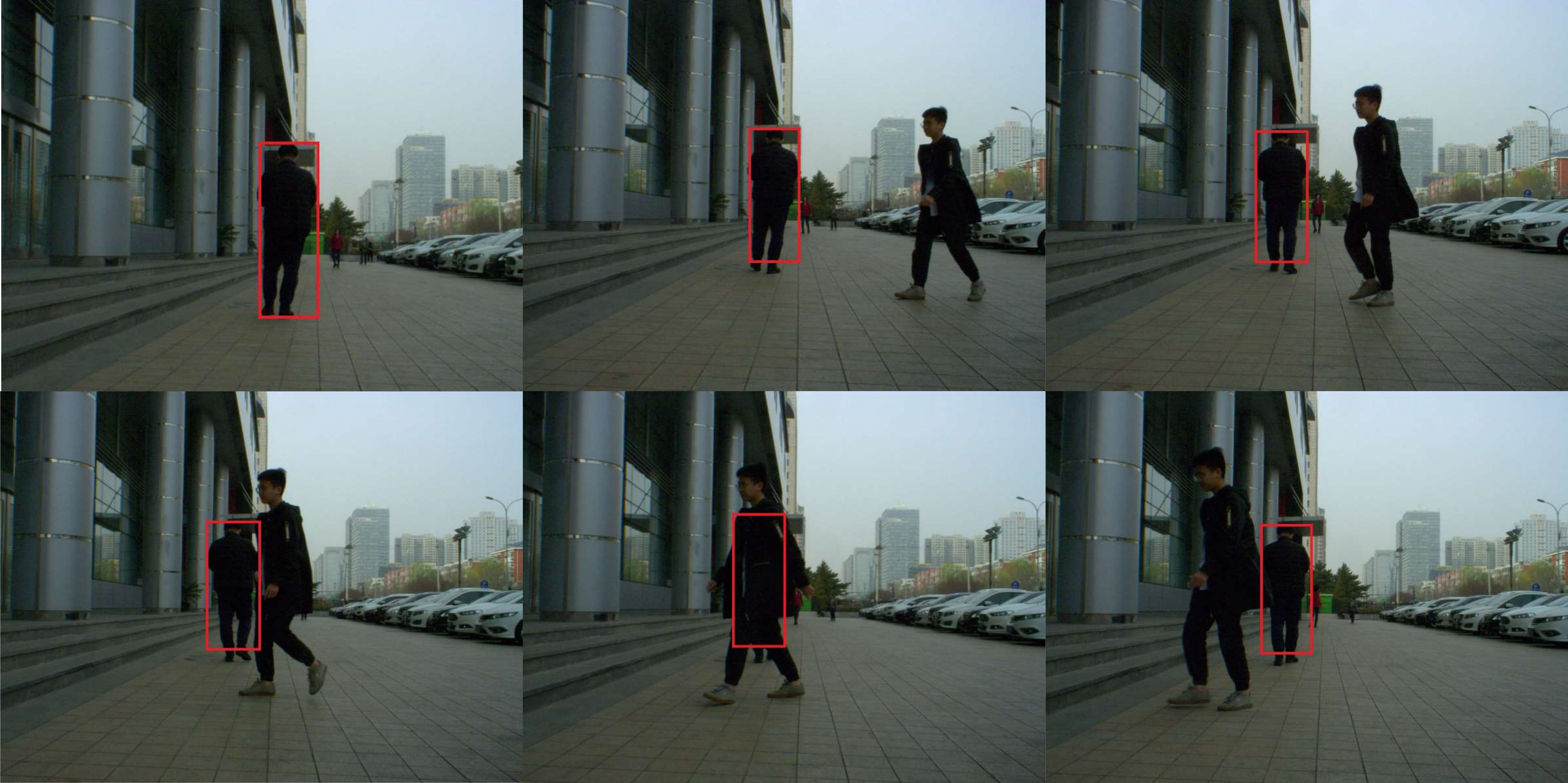}
\label{following_occlusion}
}
\subfigure[]{
\includegraphics[width=0.8\linewidth]{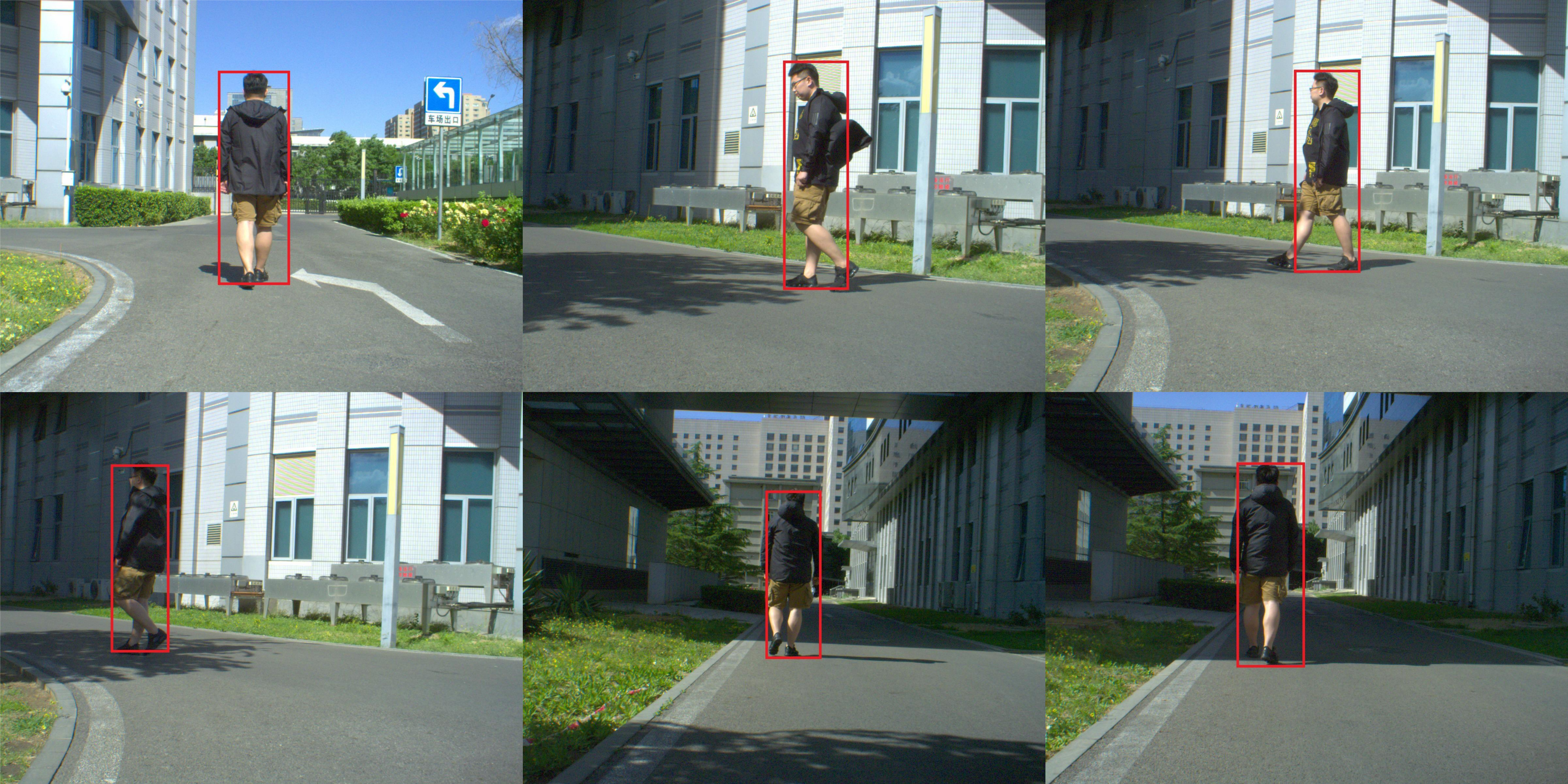}
\label{outdoor_turn}
}
\caption{\small Human tracking results in outdoor experiments. (a) Occlusion and distractor challenge. (b) Background changes challenge.}
\label{outdoor_results}
\end{figure}

Figure \ref{indoor_results} and  Figure \ref{outdoor_results} further visualize the tracking results in human following task. Due to flow aggregation and box regression modules, FlowTrack++ can handle the pose changes and scale challenges in Figure \ref{indoor_results}. In Figure \ref{outdoor_results}, targets are under occlusion, distractor and background changes. Since hard-negative samples mining and failure recovering strategies are adopted in FlowTrack++ algorithm, these challenges can be handled.

\section{Conclusion}

In this paper, we propose a human following system on mobile robot with monocular pan-tilt camera, which mainly consists of a visual tracker and a motion controller. In visual tracking algorithm, both Siamese networks and optical flow information are exploited to locate and regress human simultaneously. Besides, a motion controller is derived to stay the target in the field of view and keep following simultaneously, which does not need the depth sensors.  In experiments, the overall system obtains accurate and robust following results both in simulations and real robot platform. Future work will explore the multi-object tracking and re-identification for human following task.



\footnotesize
\bibliographystyle{IEEEtran}
\footnotesize
\bibliography{egbib}

\end{document}